\pdfoutput=1

\documentclass[11pt]{article}

\usepackage{ACL2023}

\usepackage{times}
\usepackage{latexsym}

\usepackage[T1]{fontenc}

\usepackage[utf8]{inputenc}

\usepackage{inconsolata}
\usepackage{microtype}
\usepackage{graphicx}
\usepackage{xcolor}
\usepackage{todonotes}
\usepackage{multirow}
\usepackage{amsmath,amssymb}
\usepackage{bbm}
\usepackage{array}

\usepackage{float} 
\usepackage{subfigure}
\usepackage{epstopdf}
\usepackage{tabularx}

\newcommand{\model}{\textsc{Changes}}
\newcommand\Tstrut{\rule{0pt}{2.6ex}}         
\newcommand\Bstrut{\rule[-0.9ex]{0pt}{0pt}}   

\title{Contrastive Hierarchical Discourse Graph for Scientific Document Summarization}

\author{Haopeng Zhang, Xiao Liu, Jiawei Zhang \\       IFM Lab, Department of Computer Science, University of California, Davis, CA, USA \\\texttt{haopeng,xiao,jiawei@ifmlab.org}}

\begin{document}
\maketitle
\begin{abstract}
The extended structural context has made scientific paper summarization a challenging task. This paper proposes \textsc{changes}, a contrastive hierarchical graph neural network for extractive scientific paper summarization. \textsc{changes} represents a scientific paper with a hierarchical discourse graph and learns effective sentence representations with dedicated designed hierarchical graph information aggregation. We also propose a graph contrastive learning module to learn global theme-aware sentence representations. Extensive experiments on the PubMed and arXiv benchmark datasets prove the effectiveness of \textsc{changes} and the importance of capturing hierarchical structure information in modeling scientific papers.

\end{abstract}

\section{Introduction}
Extractive document summarization aims to extract the most salient sentences from the original document and form the summary as an aggregate of these sentences. Compared to abstractive summarization approaches that suffer from hallucination generation problems~\cite{kryściński2019evaluating,zhang2022improving}, summaries generated in an extractive manner are more fluent, faithful, and grammatically accurate, but may lack coherence across sentences. Recent advances in deep neural networks and pre-trained language models~\cite{devlin2018bert,lewis2019bart} have led to significant progress in single document summarization \cite{nallapati2016summarunner,narayan2018ranking,liu2019text,zhong2020extractive}.
However, these methods mainly focus on short documents like news articles in CNN/DailyMail \cite{hermann2015teaching} and New York Times \cite{sandhaus2008new}, and struggle when dealing with relatively long documents such as scientific papers.

The challenges of lengthy scientific paper summarization lie in several aspects. First, the extended input context hinders cross-sentence relation modeling, the critical step of extractive summarization \cite{wang2020heterogeneous}. Thus, sequential models like RNN are incapable of capturing the long-distance dependency between sentences, and hard to differentiate salient sentences from others. Furthermore, scientific papers tend to cover diverse topics and contain rich hierarchical discourse structure information. The internal hierarchy structure, like sections, paragraphs, sentences, and words, is too complex for sequential models to capture. Scientific papers generally follow a standard discourse structure of problem definition, methodology, experiments and analysis, and conclusions \cite{xiao2019extractive}. Moreover, the lengthy input context also makes the widely adopted self-attention Transformer-based models \cite{vaswani2017attention} inapplicable. The input length of a scientific paper can range from $2000$ to $7,000$ words, which exceeds the input limit of the Transformer due to the quadratic computation complexity of self-attention. Thus, sparse Transformer models like BigBird~\cite{zaheer2020big} and Longformer \cite{beltagy2020longformer} are proposed.

Recently, researchers have also turned to graph neural networks (GNN) as an alternative approach. Graph neural networks have been demonstrated to be effective at tasks with rich relational structure and can preserve global
structure information \cite{yao2019graph,xu2019discourse,zhang2020text}. By representing a document as a graph, GNNs update and learn sentence representations by message passing, and turn extractive summarization into a node classification problem. Among all attempts, one popular way is to construct cross-sentence similarity graphs \cite{erkan2004lexrank,zheng2019sentence}, which uses sentence representation cosine similarity as edge weights to model cross-sentence semantic relations. \citet{xu2019discourse} proposed using Rhetorical Structure Theory (RST) trees and coreference mentions to capture cross-sentence discourse relations. \citet{wang2020heterogeneous} proposed constructing a word-document heterogeneous graph by using words as the intermediary between sentences. Despite their success, how to construct an effective graph to capture the hierarchical structure for academic papers remains an open question.

To address the above challenges, we propose {\model} (\textbf{C}ontrastive \textbf{H}ier\textbf{A}rchical \textbf{G}raph neural network for \textbf{E}xtractive \textbf{S}ummarization), a hierarchical graph neural network model to fully exploit the section structure of scientific papers. 
{\model} first constructs a sentence-section hierarchical graph for a scientific paper, and then learns \textit{hierarchical} sentence representations by dedicated designed information aggregation with iterative intra-section and inter-section message passing. Inspired by recent advances in contrastive learning \cite{liu2021simcls,chen2020simple}, we also propose a graph contrastive learning module to learn global theme-aware sentence representations and provide fine-grained discriminative information. The local sentence and global section representations are then fused for salient sentence prediction. We validate {\model} with extensive experiments and analyses on two scientific paper summarization datasets. Experimental results demonstrate the effectiveness of our proposed method. Our main contributions are as follows:

\begin{itemize}
    \item We propose a hierarchical graph-based model for long scientific paper extractive summarization. Our method utilizes the hierarchical discourse of scientific documents and learns effective sentence representations with iterative intra-section and inter-section sentence message passing.
    
    \item We propose a plug-and-play graph contrastive module to provide fine-grained discriminative information. The graph contrastive module learns global theme-aware sentence representations by pulling semantically salient neighbors together and pushing apart unimportant sentences. Note that the module could be added to any extractive summarization system.
    
    \item  We validate our proposed model on two benchmark datasets (arXiv and PubMed), and the experimental results demonstrate its effectiveness over strong baselines.

\end{itemize}


\begin{figure*}[!htbp]
\centering
\begin{minipage}[t]{\textwidth}
\centering
\includegraphics[width=\textwidth]{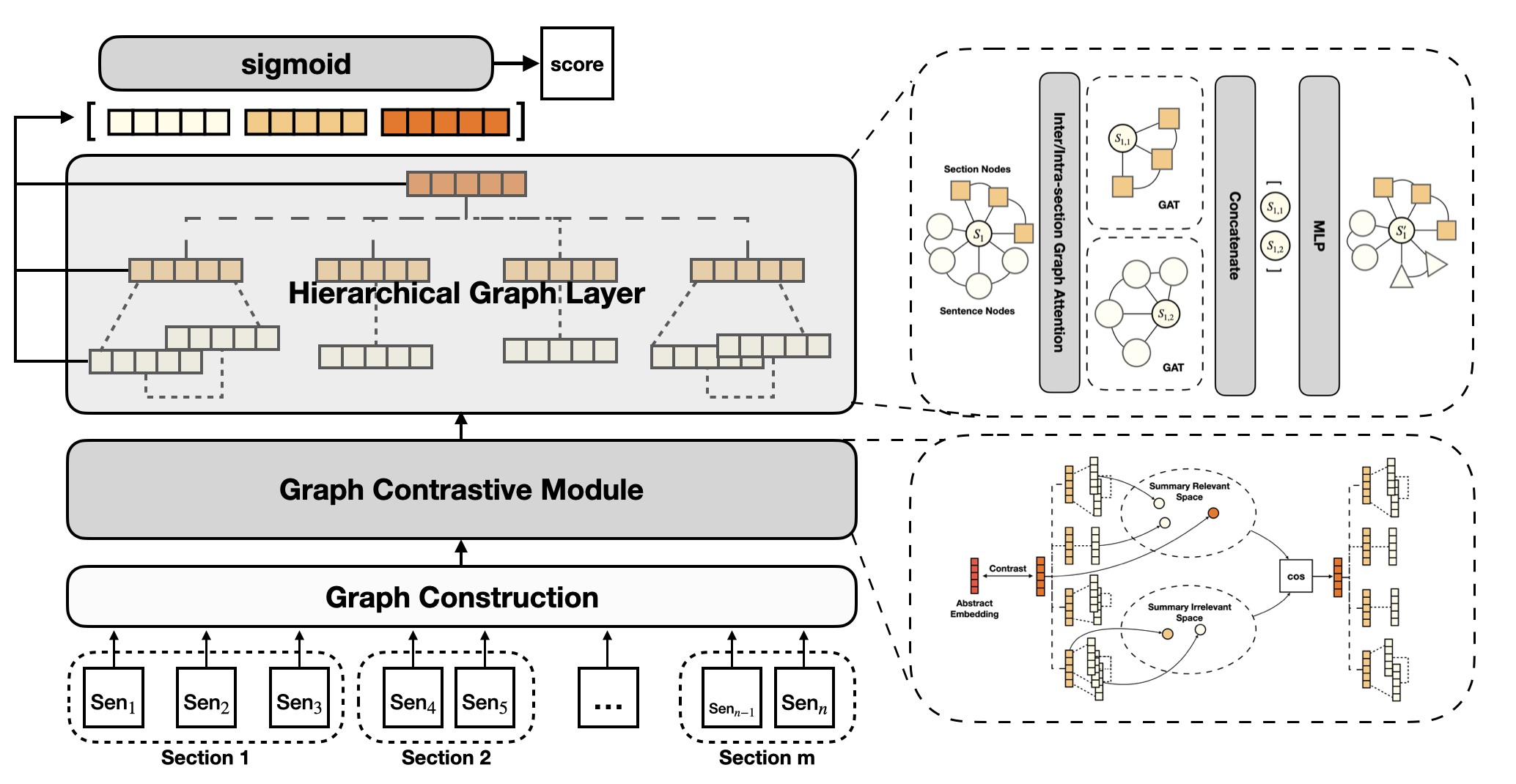}
\caption{The overall model architecture of {\model}. We first construct a hierarchical graph for an input document, and then learn representations with a graph contrastive module and hierarchical graph layers. The concatenation representations of
sentence node and its section node will be fused for summary sentence selection.}
\label{architecture}
\end{minipage}
\end{figure*}

\section{Related Work}
\subsection{Extractive Summarization on Scientific Papers}

Despite the superior performance on news summarization by recent neural network models~\cite{zhou2018neural,zhang2023diffusum,zhang2023extractive,fonseca2022factorizing} and pre-trained language models \cite{liu2019text,lewis2019bart}, progress in long document summarization such as scientific papers is still limited. 


Traditional approaches to summarize scientific articles rely on supervised machine learning algorithms such as LSTM \cite{collins2017supervised} with surface features such as sentence position, and section categories. Recently, \citet{xiao2019extractive} proposed a neural-based method by incorporating both the global context of the whole document and the local
context within the current topic with an encoder-decoder model. \citet{ju2021leveraging} designed an unsupervised extractive approach to summarize long scientific documents based on the Information Bottleneck
principle. \citet{dong2020discourse} proposed an unsupervised ranking model by incorporating two-level hierarchical graph representation and asymmetrical positional cues to determine sentence importance. Recent works also apply pre-trained sparse language models like Longformer for modeling long documents~\cite{beltagy2020longformer,ruan2022histruct+,cho2022toward}.

\subsection{Graph-based Summarization}
Graph models have been widely applied to extractive summarization due to the capability of modeling cross-sentence relations within a document. The sparsity nature of graph structure also brings scalability and flexibility, making it a good fit for long documents. Graph neural networks' memory costs are generally linear with regard to the input size compared to the quadratic self-attention mechanism.


Researchers have explored supervised graph neural network methods for summarization \cite{cui2021topic,jia2020neural,huang2021extractive,xie2022gretel,phan2022hetergraphlongsum}. \citet{yasunaga2017graph} first proposed to use Graph Convolutional Network (GCN) on the approximate discourse graph. \citet{xu2019discourse} then applied GCN on structural discourse graphs based on RST trees and coreference mentions. Recently, \citet{wang2020heterogeneous} proposed constructing a word-document heterogeneous graph by using words as the intermediary between sentences. \citet{zhang2022hegel} proposed to use hypergraph to capture the high-order sentence relations within the document. Our paper follows the series of work but incorporates hierarchical graphs for scientific paper discourse structure modeling and graph contrastive learning for theme-aware sentence representation learning.



\section{Method}
Given a document $D = \{{s}_1, {s}_2, ..., {s}_n\}$ with $n$ sentences and $m$ sections, we first represent it as a hierarchical graph and formulate extractive summarization as a node labeling task. The objective is to predict labels $y_i \in (0, 1)$ for all sentences, where $y_i = 1$ and $y_i = 0$ represent whether the $i$-th sentence should be included in the summary or not, respectively. 

The overall model architecture of {\model} is shown in Figure~\ref{architecture}. {\model} consists of two modules: a graph contrastive learning module to learn global theme-aware sentence representations and a hierarchical graph layer module to learn hierarchical graph node representations with iterative message passing. The learned sentence node and section node representations will be used as indicators for salient sentence selection.

\subsection{Graph Construction}
\begin{figure}[!htbp]
\centering
\begin{minipage}[t]{0.45\textwidth}
\centering
\includegraphics[width=\textwidth]{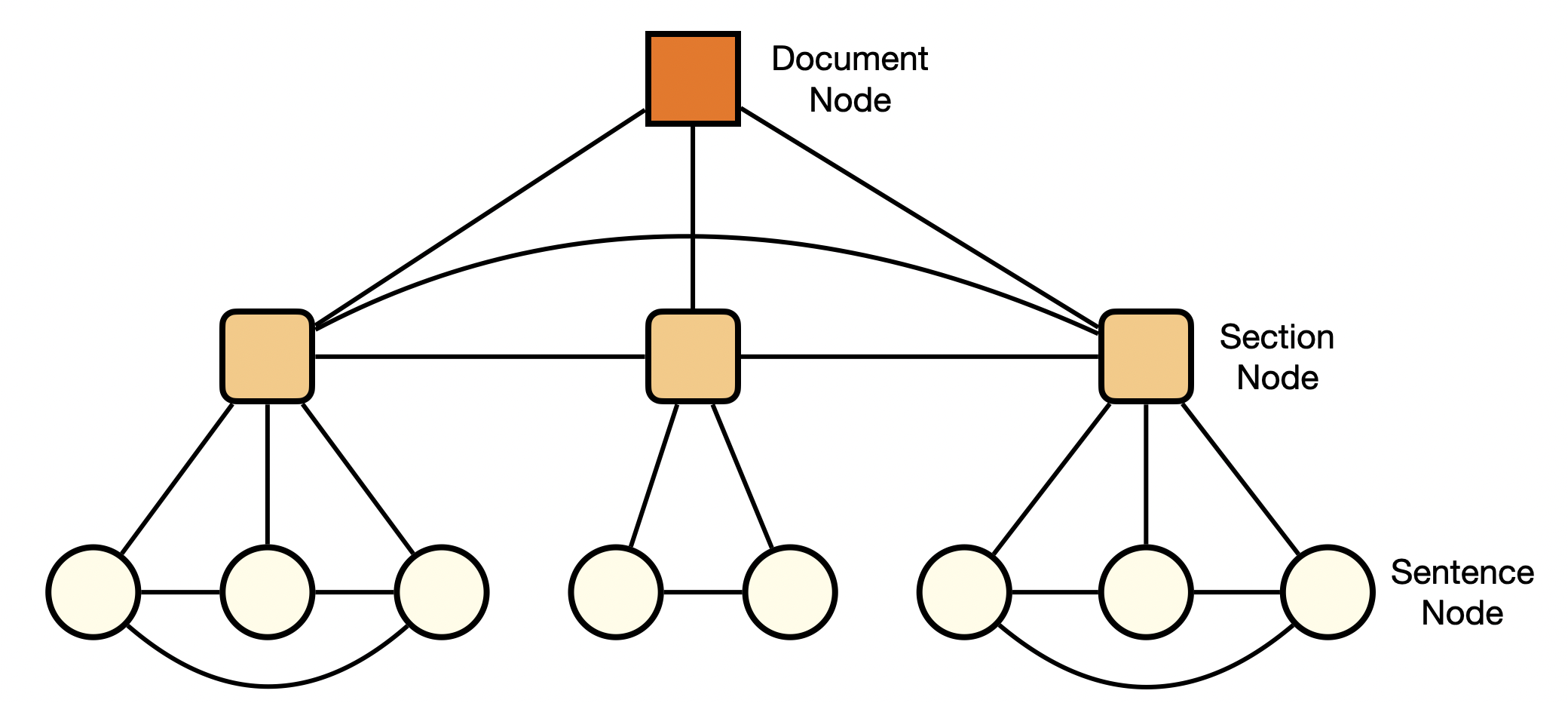}
\caption{An illustration of a hierarchical graph for a long input document with rich discourse structures.}
\label{graph}
\end{minipage}
\end{figure}

Given an academic paper $D$, we first construct a hierarchical graph $\mathcal{G} = (\mathcal{V},\mathcal{E})$, where ${\mathcal{V}}$ stands for the node set and $\mathcal{E}$ represents edges between nodes. In order to utilize the sentence-section hierarchical structure of academic papers, the undirected hierarchical graph $\mathcal{G}$ contains both sentence nodes and section nodes, defined by $\mathcal{V} =\mathcal{V}_{sen} \cup \mathcal{V}_{sec}$, where each sentence node ${v}_{{sen}_i} \in \mathcal{V}_{sen}$ represents a  sentence $s_i$ in the document $D$ and ${v}_{{sec}_j} \in \mathcal{V}_{sec}$ represents one section in the document. The edge connection of $\mathcal{G}$ is defined as $\mathcal{E} =\mathcal{E}_{sen} \cup \mathcal{E}_{sec} \cup \mathcal{E}_{cross}$, where $\mathcal{E}_{sen}$ denotes the connection between sentence nodes within the same section, $\mathcal{E}_{sec}$ denotes the connection between section nodes, and $\mathcal{E}_{cross}$ denotes the cross-connection between a sentence node and its corresponding section node. Note that we also add a special section supernode $v_D$ that represents the whole document $D$. An illustration of the hierarchical graph is shown in Figure~\ref{graph}.

\paragraph{Edge Connection}
Unlike prior work \cite{zheng2019sentence, dong2020discourse} that uses cosine similarity of sentence semantic representations as edge weights, we construct unweighted hierarchical graphs to disentangle structural information (adjacency matrix $\mathbf{A}$) from semantic information (node representation $\mathbf{H}$). In other words, connected nodes have weight $1$, and disconnected nodes have weight $0$ in the adjacency matrix $\mathbf{A}$.

Formally, sentence-level edge ${e}_{{sen}_{i,j}}$ connects sentence nodes ${v}_{{sen}_i}$ and ${v}_{{sen}_j}$ if they are within the same section, aiming to aggregate local intra-section information. All section nodes are fully connected by section-level edges ${e}_{{sec}_{p,q}}$, aiming to aggregate global inter-section information. The cross-level edge ${e}_{{cross}_{i,p}}$ connects the sentence node ${v}_{{sen}_i}$ to its corresponding section node ${v}_{{sec}_p}$, which allow message passing between sentence-level and section-level nodes.

In a hierarchical graph, a sentence node could only directly interact with local neighbor nodes within the same section, and indirectly interact with sentence nodes of other sections via section-level node connections.


\paragraph{Node Representation}

We adopt BERT (Bidirectional Encoder Representations from Transformers) \cite{devlin2018bert} as sentence encoder to embed the semantic meanings of sentences $\{{s}_1, {s}_2, ..., {s}_n\}$ as initial node representations $\mathbf{X}=\{\mathbf{x}_1, \mathbf{x}_2, ..., \mathbf{x}_n\}$. Note that BERT here is only used for initial sentence embedding, but is not updated during the training process to reduce model computing cost and increase efficiency.
 
In addition to the semantic representation of sentences, we also inject positional encoding following Transformer \cite{vaswani2017attention} to preserve the sequential order information. We apply the hierarchical position embedding by \cite{ruan2022histruct+} to model sentence positions accompanying our hierarchical graph. Specifically, the position of each sentence $s_i$ can be represented as two parts: its corresponding section index $p^{sec}_i$, and its sentence index within section $p^{sen}_i$. Formally, the hierarchical position embedding (HPE) of sentence $s_i$ can be calculated as:
\begin{equation}
    \text{HPE}(s_i) = \text{PE}(p^{sec}_i)+\text{PE}(p^{sen}_i),
\end{equation}
where $\text{PE}(\cdot)$ refers to the position encoding function in \cite{vaswani2017attention}:

\begin{align}
    \text{PE}(pos, 2i) = \sin(pos/10000^{2i/d}),\\
    \text{PE}(pos, 2i+1) = \cos(pos/10000^{2i/d}).
\end{align}
Overall, we can get the initial sentence node representations $\mathbf{H}^0_{sen} = \{\mathbf{h}_{sen_1}^0, \mathbf{h}_{sen_2}^0, ..., \mathbf{h}_{sen_n}^0\}$, with vector $\mathbf{h}_i^0\in \mathbbm{R}^{d}$ defined as:
\begin{equation}
    \mathbf{h}_i^0 = \mathbf{x}_i + \text{HPE}(s_i),
\end{equation}
where $d$ is the dimension of the node embedding. The initial section node representation $\mathbf{h}_{sec_j}^0\in \mathbbm{R}^{d}$ for the $j$-th section is the mean of its connected sentences embeddings, and the document node representation $\mathbf{h}_{doc}^0\in \mathbbm{R}^{d}$ is the mean of all section node embeddings.


\subsection{Graph Contrastive Module}

After constructing the hierarchical graph with adjacency matrix $\mathbf{A}$ and node representation $\mathbf{H}_{sen}^0 \in \mathbbm{R}^{n\times d} $, we apply a graph contrastive learning (GCL) module to capture global context information. Motivated by the principal idea that a good summary sentence should be more semantically similar to the source document than the unqualified sentences \cite{radev2004centroid,zhong2020extractive}, our GCL module updates sentence representations using Graph Attention Network \cite{velivckovic2017graph} with a contrastive objective to learn the global theme-aware sentence representations. Note that the module could be added to any extractive summarization system. 


\paragraph{Graph Attention Network}
Given a constructed graph $\mathcal{G} = (\mathcal{V},\mathcal{E})$ with
node representations $\mathbf{H}$ and adjacent matrix $\mathbf{A}$, a GAT layer updates a node $v_i$ with representation $\mathbf{h}_{i}$ by:

\begin{equation}
\begin{aligned}
e_{i j} &=\operatorname{LeakyReLU}\left(\mathbf{W}_a\left[\mathbf{W}_{in} \mathbf{h}_{i} \| \mathbf{W}_{in} \mathbf{h}_{j}\right]\right), \\
\alpha_{i j} &=\frac{\exp \left(e_{i j}\right)}{\sum_{l \in \mathcal{N}_{i}} \exp \left(e_{i l}\right)}, \\
\mathbf{h}_{i}^{\prime} &=\sigma\left(\sum_{j \in \mathcal{N}_{i}} \alpha_{i j} \mathbf{W}_v \mathbf{h}_{j}\right),
\end{aligned}
\end{equation}
where $\mathcal{N}_{i}$ denotes the $1$-hop neighbors of node $v_i$, $\alpha_{i j}$ denotes the attention weight between nodes $\mathbf{h}_{i}$ and $\mathbf{h}_{j}$, $\mathbf{W}_{in},\mathbf{W}_a,\mathbf{W}_v$ are trainable weight matrices, and $\|$ denotes concatenation operation.

The above single-head graph attention is further extended to multi-head attention, where $T$ independent attention mechanisms
are conducted and their outputs are concatenated as:

\begin{equation}
\mathbf{h}_{i}^{\prime}=\|_{t=1}^{T} \sigma\left(\sum_{j \in \mathcal{N}_{i}} \alpha_{i j}^{t} \mathbf{W}_{h}^{k} \mathbf{h}_{j}\right)
\end{equation}

\paragraph{Contrastive Loss}
Contrastive learning aims to learn effective representation by pulling semantically close neighbors
together and pushing apart non-neighbors \cite{marelli2014sick}. Recent works have demonstrated contrastive learning to be effective in high-order representation learning \cite{chen2020simple,gao2021simcse}. Thus, we optimize our GCL module in a contrastive manner with the following contrastive loss. The goal of contrastive learning is to learn theme-aware sentence embedding by pulling semantically salient neighbors together and pushing apart less salient sentences. The contrastive loss is formally defined as:
\begin{equation}
\begin{aligned}
\mathcal{L}_c &= -\log \frac{\exp(sim((\mathbf{h}_{D}^{\prime}, \mathbf{h}_i^{\prime})/\tau)}{\sum_{j=1}^{n} \exp(sim(\mathbf{h}_{D}^{\prime}, \mathbf{h}_j^{\prime}) / \tau)},
\end{aligned}
\label{cont}
\end{equation}
where $\mathbf{h}_{D}^{\prime}$ is the document node embedding, $\mathbf{h}_i^{\prime}$ is the updated representaion of sentence $s_i$ , and $\tau$ is the temperature factor.

After passing through the GCL module, the learned global theme-aware sentence embeddings $\mathbf{H}^c_{sen} = \{\mathbf{h}_{sen_1}^c, \mathbf{h}_{sen_2}^c, ..., \mathbf{h}_{sen_n}^c\} \in \mathbbm{R}^{n \times d}$ are then passed to the hierarchical graph layer module.

\subsection{Hierarchical Graph Layer}
To exploit the sentence structure of academic papers, {\model} then updates sentence and section node representations with hierarchical graph layers in an iterative manner.

The hierarchical graph layer first updates sentence embeddings with the \text{local} neighbor sentences within the same section with GAT for intra-section message passing, then update section nodes with sentence nodes for cross-level information aggregation to exploit the hierarchical structure of academic papers. Next, inter-section message passing allow \textit{global} context information interaction. Finally, the sentence nodes are updated based on their corresponding section node, fusing both local and global context information.

Formally, each iteration contains four update processes: one intra-section message passing, one sentence-to-section aggregation, one inter-section message passing, and finally one section-to-sentence aggregation. For the $l$-th iteration, the process can be represented as:




\begin{equation}
  \begin{aligned}
    \mathbf{H}_{sen}^{\prime} &= \text{GAT}(\mathbf{H}_{sen}^l)\\
    \mathbf{H}_{sec}^{\prime} &= \text{GAT}(\mathbf{H}_{sen}^{\prime})\\
        \mathbf{H}_{sec}^{l+1} &= \text{GAT}(\mathbf{H}_{sec}^{\prime})\\
        \mathbf{H}_{sen}^{l+1} &= \sigma({\mathbf{W}_{b}}[\mathbf{H}_{sen}^{\prime}\|\mathbf{H}_{sec}^{l+1}])
  \end{aligned}
\end{equation}

where $\mathbf{H}_{sen}^{\prime},\mathbf{H}_{sec}^{\prime}$ denotes the intermediate representations of sentence and section nodes, $\mathbf{H}_{sen}^{l+1},\mathbf{H}_{sec}^{l+1}$ denotes the updated sentence and section node representations, and $[\mathbf{H}_{sen}^{\prime}\|\mathbf{H}_{sec}^{l+1}]$ denotes the concatenation of intermediate sentence node representation and its corresponding updated section node representation.

In this way, {\model} updates and learns hierarchy-aware sentence embeddings through the hierarchical graph layers.





\begin{table}
\small
\centering
\begin{tabular}{c |c |c }

\hline & \text { Arxiv } & \text { PubMed } \\
\hline
\# {train} & 201,427 & 112,291 \\
\# {validation} & 6,431 & 6,402 \\
\# {test} & 6,436 & 6,449 \\
\text {avg. word/doc} & 4,938  & 3,016  \\
\text {avg. word/summary} & 203  & 220  \\
\text {avg. sent./doc} & 205  & 140   \\
\text {avg. sent./summary} & 5  &  6 \\
\hline
\end{tabular}
\caption{Statistics of PubMed and Arxiv datasets.}
\label{stat}
\end{table}

\subsection{Optimization}

After passing $L$ hierarchical graph layers, we obtain the final sentence node representations $\mathbf{H}^L_{sen} = \{\mathbf{h}_{sen_1}^L, \mathbf{h}_{sen_2}^L, ..., \mathbf{h}_{sen_n}^L\} \in \mathbbm{R}^{n \times d}$. We then add a multi-layer perceptron (MLP) followed by a sigmoid activation
function to indicate the confidence scores for extracting each sentence in the summary. 

Formally, the predicted confidence score ${\hat{y}}_i$ to extract a sentence $s_i$ in section ${sec_j}$ as a summary sentence is: 
\begin{equation}
       \mathbf{z}_i  = \text{LeakyReLU}({\mathbf{W}_{o1}}[{\mathbf{h}_{sen_{i}}^L}\|{\mathbf{h}_{sec_{j}}^L}]), 
\end{equation}
\begin{equation}
            {\hat{y}}_i = \text{sigmoid}(\mathbf{W}_{o2}\mathbf{z}_i),
\end{equation}
where $\mathbf{W}_{o1},\mathbf{W}_{o2}$ are both trainable parameters, and $[{\mathbf{h}_{sen_{i}}^L}\|{\mathbf{h}_{sec_{j}}^L}]$ denote the concatenation of sentence embedding and its corresponding section embedding. During the inference phase, we will select the $k$ sentences with the highest predicted confidence scores as the extractive summary for the input long document.

Since the extractive ground truth labels for long documents are highly imbalanced, we optimize hierarchical graph layers using weighted cross entropy loss following \cite{xiao2019extractive} as:

\small
\begin{equation}
    \mathcal{L}_s =- \frac{1}{N N_d}\sum_{d=1}^{N} \sum_{i=1}^{N_{d}}(\eta \cdot y_{i} \log {\hat{y}}_i +(1-y_{i}) \log (1-{\hat{y}}_i)),
\end{equation}
\normalsize
where $N$ denotes the number of training instances in the training set, $N_{d}$ denotes the number of sentences in the document, $\eta=\frac{\#negative}{\#positive}$ denote the ratio of the number of negative and positive sentences in the document, and $y_i$ represent the ground-truth of sentence $i$. 


\paragraph{Training Loss}
Overall, we optimize {\model} in an end-to-end manner, by optimizing the graph contrastive module and hierarchical graph layers simultaneously.

The overall training loss of {\model} is: 
\begin{equation}
\begin{aligned}
\mathcal{L} &= \mathcal{L}_s + \lambda \mathcal{L}_c,
\end{aligned}
\end{equation}
where $\lambda$ is a re-scale hyperparameter and $\mathcal{L}_c$ denotes the contrastive loss in Equation~\ref{cont}.
\begin{table*}[htbp]
\centering
\small
\begin{tabular}{c | c  c  c| c  c  c}
    \hline & \multicolumn{3}{c|}{{PubMed}} & \multicolumn{3}{c}{{ArXiv}} \Tstrut\Bstrut\\
    \hline 
    {Model} & {ROUGE-1} & {ROUGE-2} &{ROUGE-L}   & {ROUGE-1} & {ROUGE-2} & {ROUGE-L} \Tstrut\Bstrut\\
    \hline 
        Oracle(15k tok.) & 53.04 & 29.08 & 48.31  & 53.58 & 26.19 & 47.76\Tstrut\\
        Lead-10 & 38.59 & 13.05 & 34.81 & 37.37 & 10.85 & 33.17 \\
    LexRank (2004) & 39.19  & 13.89 & 34.59 & 33.85  & 10.73 & 28.99 \\
    SumBasic (2007) & 37.15 & 11.36 & 33.43 & 29.47 & 6.95 & 26.30 \\
    PACSUM (2019) & 39.79 & 14.00 & 36.09 & 38.57 & 10.93 & 34.33 \\
    HIPORANK (2021) & 43.58 & 17.00 & 39.31 & 39.34 & 12.56 & 34.89 \Bstrut\\

    \hline 
    
    Cheng\&Lapata (2016) & 43.89 & 18.53 & 30.17 & 42.24 & 15.97 & 27.88 \Tstrut \\
    SummaRuNNer (2016) & 43.89 & 18.78 & 30.36 & 42.81 & 16.52 & 28.23 \\
    ExtSum-LG (2019) & 44.85 & 19.70 & 31.43 & 43.62 & 17.36 & 29.14 \\
    SentCLF (2020) & 45.01 & 19.91 & 41.16 & 34.01 & 8.71 & 30.41\\
    SentPTR (2020) & 43.30 & 17.92 & 39.47 & 42.32 & 15.63 & 38.06 \\
    ExtSum-LG + RdLoss (2021) & 45.30 & 20.42 & 40.95 & 44.01 & 17.79 & 39.09\\
    ExtSum-LG + MMR (2021) & 45.39 & 20.37 & 40.99 & 43.87 & 17.50 & 38.97 \Bstrut\\

    \hline PGN (2017) & 35.86 & 10.22 & 29.69 & 32.06 & 9.04 & 25.16 \Tstrut \\
    DiscourseAware (2018) & 38.93 & 15.37 & 35.21 & 35.80 & 11.05 & 31.80\\
    TLM-I+E (2020) & 42.13 & 16.27 & 39.21  & 41.62 & 14.69 & 38.03\\
    
    \hline 
    \textbf{CHANGES} (ours) & \textbf{46.43} & \textbf{21.17} & \textbf{41.58} & \textbf{45.61} & \textbf{18.02} & \textbf{40.06}\Tstrut\Bstrut\\
    \hline
\end{tabular}
\caption{ROUGE F1 results on PubMed and Arxiv datasets. We keep the same train/validation/test splitting in all the experiments and report ROUGE scores from the original papers if available, or scores from \cite{xiao2019extractive} otherwise.}
\label{exp_result}
\end{table*}
\section{Experiment}
\subsection{Experiment Setup}
\paragraph{Dataset} To validate the effectiveness of {\model}, we conduct extensive experiments on two benchmark datasets: arXiv and PubMed \cite{cohan2018discourse}. The arXiv dataset contains papers in scientific domains, while the PubMed dataset contains scientific papers from the biomedical domain. These two benchmark datasets are widely adopted in long document summarization research and we use the original train, validation, and testing splits as in \cite{cohan2018discourse}. The detailed statistics of datasets are shown in Table~\ref{stat}.

\paragraph{Evaluation} Following the common setting, we use ROUGE F-scores \cite{lin2004rouge} as the automatic evaluation metrics. Specifically, we report the ROUGE-1/2 scores to measure summary informativeness and ROUGE-L scores to measure summary fluency.
Following prior work \cite{liu2019text,nallapati2016abstractive}, we also construct extractive ground truth labels (ORACLE) for training by greedily optimizing the ROUGE score on gold-reference abstracts.


\subsection{Implementation Details}
We use the publicly released BERT-base \footnote{https://github.com/google-research/bert} \cite{devlin2018bert} as the sentence encoder. The BERT encoder is only used to generate initial sentence embeddings, but is not updated during training to improve model efficiency. We adopt the Graph Attention Network \footnote{https://github.com/PetarV-/GAT} \cite{velivckovic2017graph} implementation with $8$ attention heads and $2$ stack layers for graph message passing. The hidden size of our model is set to $2048$. 

Our model is trained with the Adam optimizer \cite{loshchilov2017decoupled} with a learning rate of 0.0001 and a dropout rate of 0.1. We train our model on a single RTX A6000 GPU for 10 epochs and validate after each epoch using ROUGE-1 F-score. We employ early stopping to select the best model for a patient duration of 3 epochs. We searched the training loss re-scale factor $\lambda$ in the range of $0$ to $1$ with $0.1$ step size and got the best value of $0.5$. 


\subsection{Baseline Methods}
We perform a systematic comparison with recent approaches in both extractive and abstractive summarization for completeness. We keep the same train/validation/test splitting in all the experiments and report ROUGE scores from the original papers if available, or scores from \cite{xiao2019extractive} otherwise. Specifically, we compare with the following strong baseline approaches:

\noindent
\textbf{Unsupervised methods}: LEAD method that selects the first few sentences as a summary, SumBasic \cite{vanderwende2007beyond}, graph-based unsupervised models LexRank \cite{erkan2004lexrank}, PACSUM \cite{zheng2019sentence} and HIPORANK \cite{dong2020discourse}.

\noindent
\textbf{Neural extractive models}: encoder-decoder based model Cheng\&Lapata \cite{cheng2016neural} and SummaRuNNer \cite{nallapati2016summarunner}; local and global context model ExtSum-LG \cite{xiao2019extractive} and its variants ExtSum-LG+RdLoss/MMR \cite{xiao2020systematically}; language model-based methods SentCLF and SentPTR \cite{subramanian2019extractive}.

\noindent
\textbf{Neural abstractive models}: pointer network generation model PGN \cite{see2017get}, hierarchical attention generation model DiscourseAware \cite{cohan2018discourse}, and transformer-based generation model {TLM-I+E} \cite{subramanian2019extractive}.

\subsection{Experiment Results}

Table~\ref{exp_result} shows the performance comparison of {\model} and all baseline methods on both PubMed and arXiv datasets. 
The first blocks include the extractive ground truth ORACLE, position-based sentence selection method LEAD, and other unsupervised baseline approaches. The second block covers state-of-the-art supervised extractive neural baselines, and the third block covers the supervised abstractive baselines.

According to the results, HIPORANK \cite{dong2020discourse} achieves state-of-the-art performance for graph-based unsupervised methods. Compared to PACSUM \cite{zheng2019sentence}, the only difference is that HIPORANK incorporates section structural information for degree centrality calculation. The performance gain demonstrates the significance of capturing the hierarchical structure of academic papers when modeling cross-sentence relations.

Interestingly, the LEAD approach performs far better when summarizing short news like CNN/DailyMail \cite{hermann2015teaching} and New York Times \cite{sandhaus2008new} than summarizing academic papers, as shown in Table~\ref{exp_result}. The results show that the distribution of ground truth sentences in academic papers is more even. In other words, academic papers have less positional bias than news articles.

We also notice that the neural extractive models tend to outperform the neural abstractive methods in general, possibly because the extended context is more challenging for generative models during decoding. ExtSum-LG \cite{xiao2019extractive} is a benchmarked extractive method with section information by incorporating both the global context of the whole document and the local context within the current topic. We argue that {\model} could better model the complex sentence structural information with the hierarchical graph than the LSTM-minus in ExtSum-LG. 

According to the experimental results, our model {\model} outperforms
all baseline approaches significantly in terms of ROUGE F1 scores on both
PubMed and arXiv datasets. The performance improvements demonstrate the usefulness of the global theme-aware representations from the graph contrastive learning module and the hierarchical graph structure for identifying the salient sentences.

\begin{figure*}[htbp]
\centering
\subfigure[ArXiv]{\includegraphics[width=0.48\textwidth]{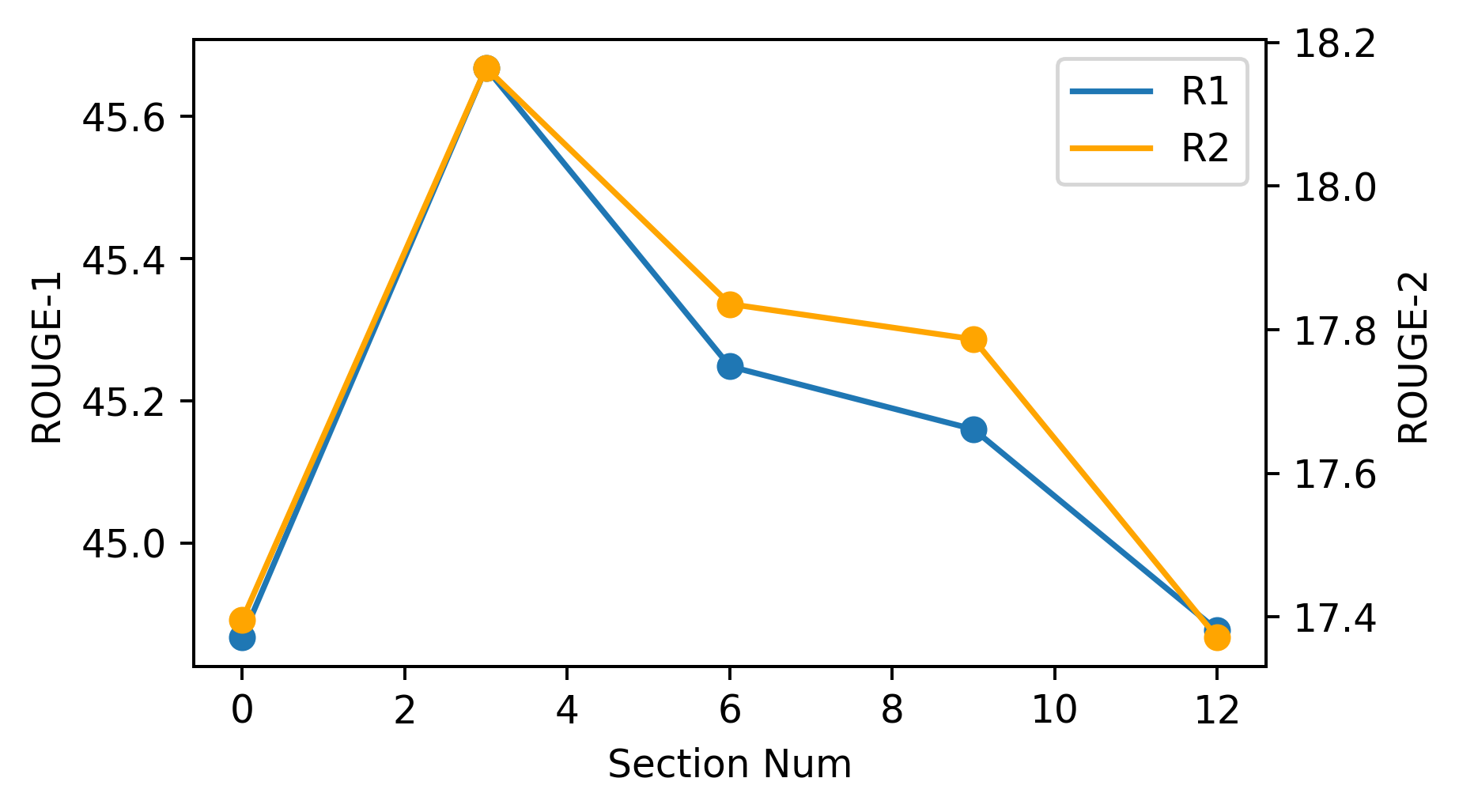}}
\subfigure[PubMed]{\includegraphics[width=0.48\textwidth]{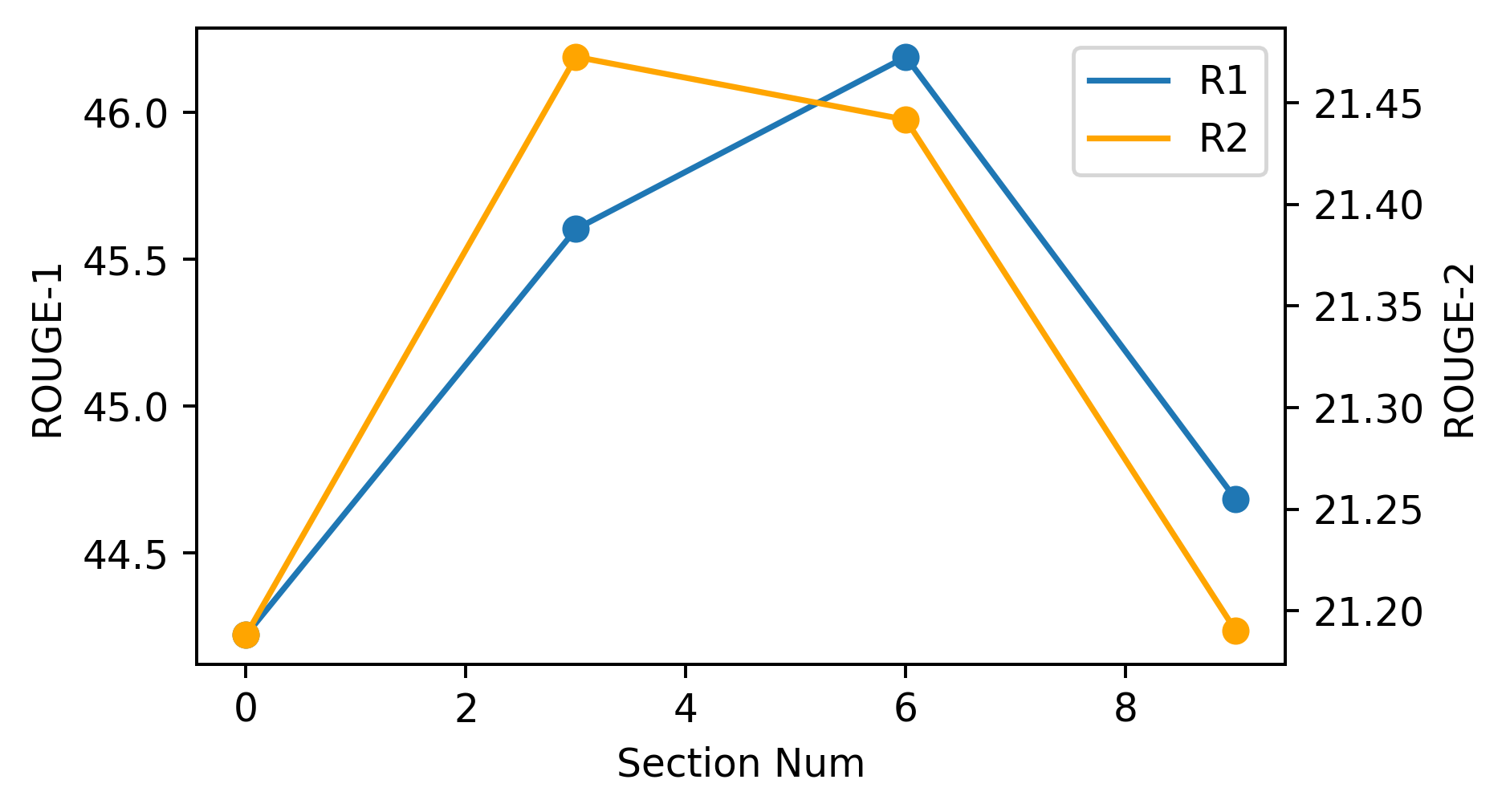}}
\caption{ROUGE-1,2 performance of {\model} for test papers with different section numbers.}
\label{rouge_sec}
\end{figure*}
\section{Analysis}
\begin{table}[htbp]
\centering
\small
\begin{tabular}{c| c c c}
    \hline Model & ROUGE-1 & ROUGE-2 & ROUGE-L \Tstrut\Bstrut\\\hline
    \multicolumn{4}{c}{PubMed}\\
    \hline 
 \textbf{\model} & \textbf{46.43} & \textbf{21.17} & \textbf{41.58}  \\
     w/o GCL & 43.91  & 18.57 & 40.01 \\
     w/o Hierarchical & 43.76 & 18.30 & 39.88 \\
    \hline
    \multicolumn{4}{c}{arXiv}\\
        \hline 
        \textbf{\model} & \textbf{45.61} & \textbf{18.02} & \textbf{40.06}\\
     w/o GCL & 44.47  & 16.58 & 38.87 \\

     w/o Hierarchical & 44.72 & 16.79 & 39.10 \\
     \hline
\end{tabular}
\caption{Ablation study results of removing components of {\model} on PubMed and arXiv datasets.}
\label{abla}
\end{table}

\begin{figure*}[htbp]
\centering
\subfigure[ArXiv]{\includegraphics[width=0.48\textwidth]{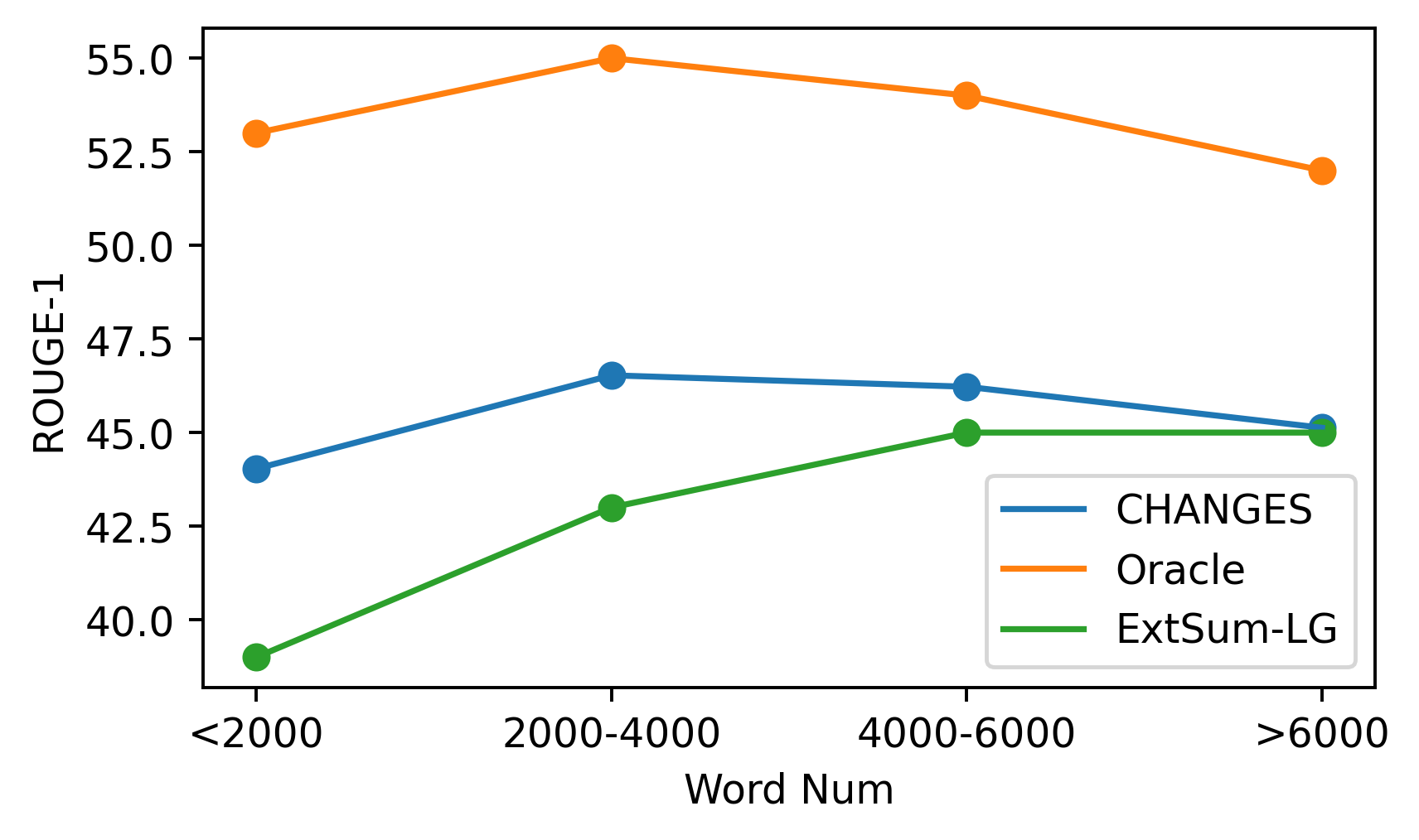}}
\subfigure[PubMed]{\includegraphics[width=0.48\textwidth]{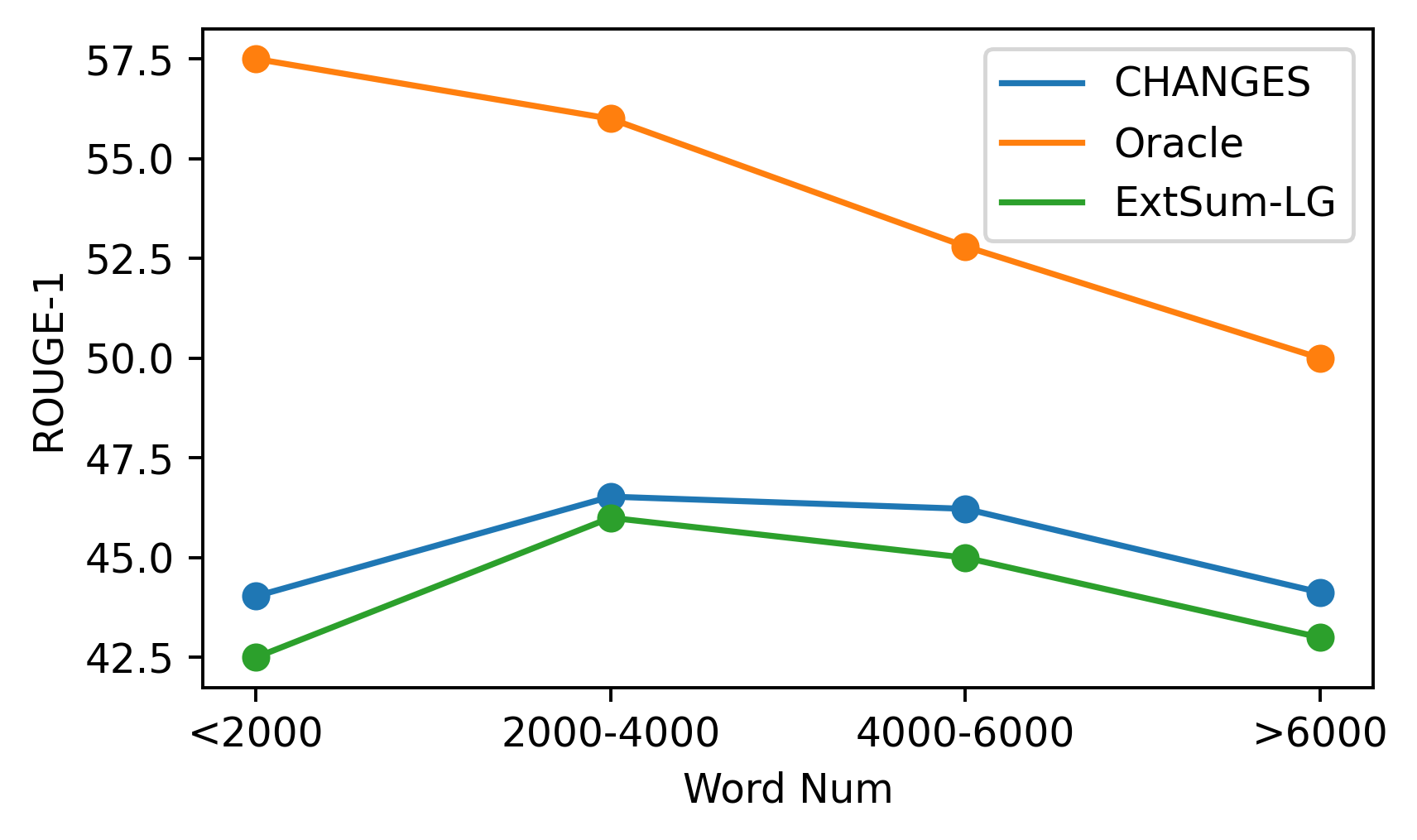}}
\caption{ROUGE-1 performance of ExtSum-LG, {\model}, ORACLE for test papers with different lengths.}
\label{rouge_word}
\end{figure*}

\subsection{Ablation Study}
\label{sec:abl}
We first analyze the influence of different components of {\model} in Table~\ref{abla}. Here the second row 'w/o Contra' means we remove the GCL module and do not update the theme-aware sentence embeddings. The third row 'w/o Hierarchical' denotes that we only use the theme-aware sentence embedding for prediction without hierarchical graph layers. As shown in the table, removing either component causes a significant model performance drop, which indicates that modeling sequential order information, semantic information, and hierarchical structural information are all necessary for academic paper summarization.

Interestingly, the theme-aware sentence embeddings and the hierarchy structure-aware sentence embeddings are almost equally critical to sentence salience modeling. The finding indicates the importance of modeling cross-sentence relations from both semantic and discourse structural perspectives.

\subsection{Performance Analysis}

We also analyze the sensitivity of {\model} to section structure and length of academic papers.
As shown in Figure~\ref{rouge_sec}, we see a performance drop trending when the number of sections increases. This is likely because the complex section structure hinders the inter-section sentence interactions. The model performance on the arXiv dataset is more stable compared to the PubMed dataset although documents in the arXiv dataset are relatively longer. We notice the same trend in Figure~\ref{rouge_word}, model performance is also more stable on arXiv datasets across different document lengths. We argue this may imply that our model is more fit for longer documents that have richer discourse structures.

Regarding the document length, we see a steady performance gain when comparing to benchmark baseline methods ExtSum-LG on both datasets as shown in Figure~\ref{rouge_word}. We also see as the document length increases, the performance gap between {\model} and extractive summary performance ceiling ORACLE becomes smaller. The finding also verifies that {\model} is especially effective and fit for long academic papers modeling. 
\section{Conclusion}
In this paper, we propose {\model}, a contrastive hierarchical graph-based model for scientific paper extractive summarization. {\model} first leans global theme-aware sentence representations by graph contrastive learning module. Moreover, {\model} incorporates the sentence-section hierarchical structure by separating intra-section and inter-section message passing and aggregating both global and local information for effective sentence embedding. Automatic evaluation on the PubMed and arXiv benchmark datasets proves the effectiveness of {\model} and the importance of capturing both semantic and discourse structure information in modeling scientific papers.

In spite of the strong zero-shot performance of large language models like ChatGPT on various downstream tasks, long document modeling is still a challenging problem in the LLM era. Transformer-based GPT-like systems still suffer from the attention computing complexity problem and  will benefit from effective and efficient modeling of long documents.

\section*{Limitations}
In spite of the strong performance of {\model}, its design still has the following limitations. First, {\model} only extracts the sentence-section-document hierarchical structure of academic papers. We believe the model performance could be further improved by incorporating document hierarchy of different granularity like dependency parsing trees and Rhetorical structure theory trees. We leave this for future work. In addition, we only focus on single academic paper summarization in this work. Academic papers generally contain a large amount of domain knowledge, thus introducing domain knowledge from peer papers or citation networks should further boost model performance.  

\section*{Acknowledgment}

This work is supported by NSF through grants IIS-1763365 and IIS-2106972. We also thank the anonymous reviewers for their helpful feedback.

\newpage
\bibliography{custom}
\bibliographystyle{acl_natbib}



\end{document}